\documentclass[11pt]{article}
\usepackage{times}
\usepackage{authblk}
\usepackage{latexsym}
\usepackage{amsmath}
\usepackage{amssymb}
\usepackage{graphicx}
\usepackage{rotating}
\usepackage{verbatim}
\usepackage{url}
\usepackage{subcaption} 
\usepackage{subfiles}
\usepackage{verbatim}
\usepackage[]{naacl2021}

\makeatletter
\def\thanks#1{\protected@xdef\@thanks{\@thanks
        \protect\footnotetext{#1}}}
\makeatother

\title{Does BERT Pretrained on Clinical Notes Reveal Sensitive Data?}

\author[$\star$ $\Psi$ $\Upsilon$ 1]{Eric Lehman}
\author[$\star$ $\Upsilon$ 2]{Sarthak Jain\thanks{$\star$ equal contribution.}}
\author[$\Phi$]{Karl Pichotta}
\author[$\Omega$]{Yoav Goldberg}
\author[$\Upsilon$]{Byron C. Wallace}
\affil[$\Psi$]{MIT CSAIL}
\affil[$\Upsilon$]{Northeastern University}
\affil[$\Phi$]{Memorial Sloan Kettering Cancer Center}
\affil[$\Omega$]{Bar Ilan University / Ramat Gan, Israel; Allen Institute for Artificial Intelligence}
\affil[1]{\texttt{lehmer16@mit.edu}} \affil[2]{\texttt{jain.sar@northeastern.edu}}

\begin{document}
\maketitle
\begin{abstract}

Large Transformers pretrained over clinical notes from Electronic Health Records (EHR) have afforded substantial gains in performance on predictive clinical tasks.
The cost of training such models (and the necessity of data access to do so) coupled with their utility motivates parameter sharing, i.e., the release of pretrained models such as ClinicalBERT \cite{Alsentzer2019PubliclyAC}. 
While most efforts have used deidentified EHR, many researchers have access to large sets of sensitive, non-deidentified EHR with which they might train a BERT model (or similar). 
Would it be safe to release the weights of such a model if they did? 
In this work, we design a battery of approaches intended to recover Personal Health Information (PHI) from a trained BERT. 
Specifically, we attempt to recover patient names and conditions with which they are associated. 
We find that simple probing methods are not able to meaningfully extract sensitive information from BERT trained over the MIMIC-III corpus of EHR. However, more sophisticated ``attacks" may succeed in doing so: To facilitate such research, we make our experimental setup and baseline probing models available.\footnote{\url{https://github.com/elehman16/exposing_patient_data_release}.}

\end{abstract}

\section{Introduction}

Pretraining large (masked) language models such as BERT \cite{Devlin2019BERTPO} over domain specific corpora has yielded consistent performance gains across a broad range of tasks. 
In biomedical NLP, this has often meant pretraining models over collections of Electronic Health Records (EHRs) \citep{Alsentzer2019PubliclyAC}.
For example, \citet{Huang2019ClinicalBERTMC} showed that pretraining models over EHR data improves performance on clinical predictive tasks. 
Given their empirical utility, and the fact that pretraining large networks requires a nontrivial amount of compute, there is a natural desire to share the model parameters for use by other researchers in the community.

However, in the context of pretraining models over patient EHR, this poses unique potential privacy concerns: Might the parameters of trained models \emph{leak} sensitive patient information? 
In the United States, the Health Insurance Portability and Accountability Act (HIPAA) prohibits the sharing of such text if it contains any reference to Protected Health Information (PHI). 
If one removes all reference to PHI, the data is considered ``deidentified", and is therefore legal to share.

\begin{figure*}
    \centering
    \includegraphics[width=0.55\textwidth]{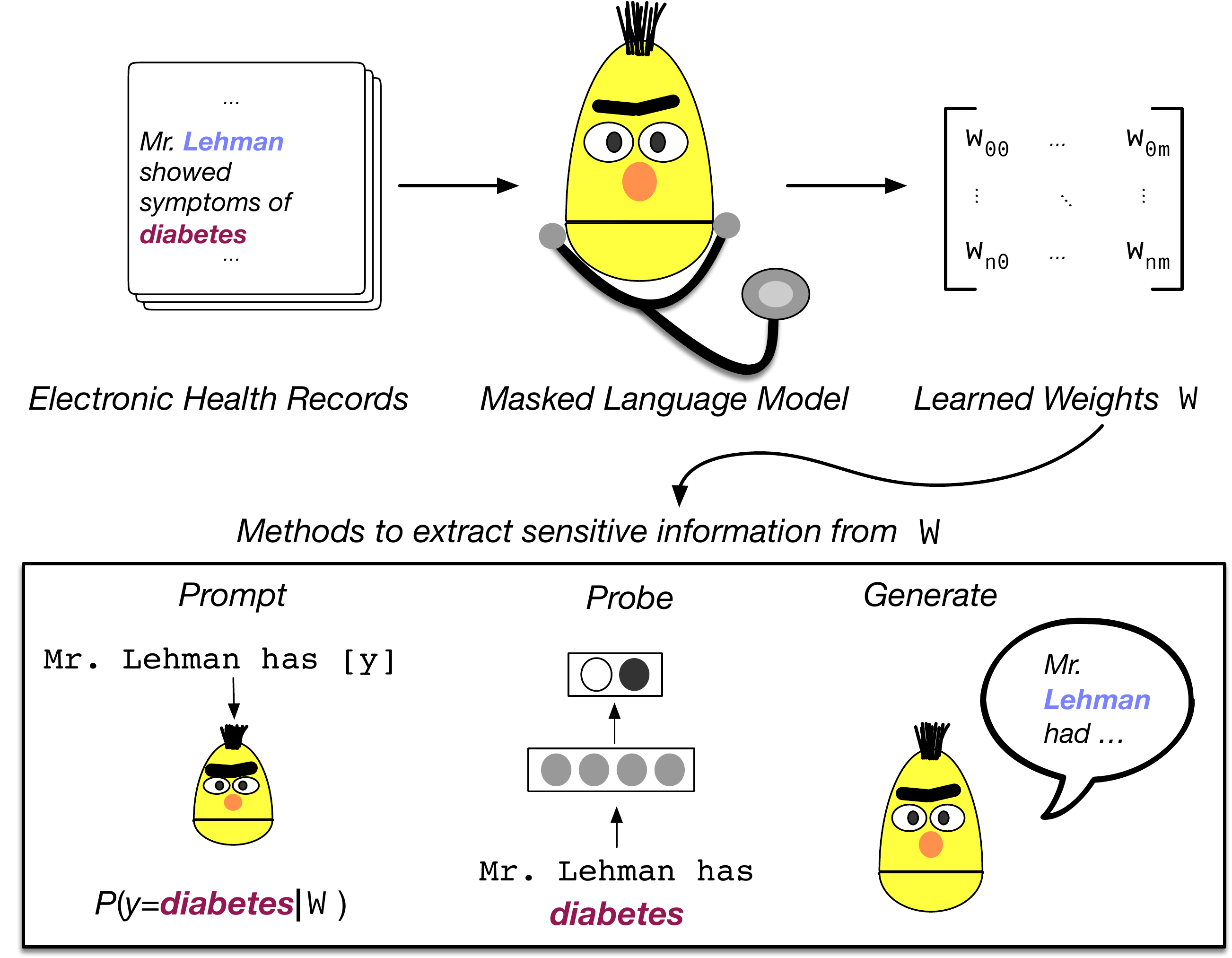}
    \caption{Overview of this work. We explore initial strategies intended to extract sensitive information from BERT model weights estimated over the notes in Electronic Health Records (EHR) data.}
    \label{fig:overview}
\end{figure*}

While researchers may not directly share non-deidentified text,\footnote{Even for deidentified data such as MIMIC \cite{mimiciii}, one typically must complete a set of trainings before accessing the data, whereas model parameters are typically shared publicly, without any such requirement.} it is unclear to what extent \emph{models} pretrained on non-deidentified data pose privacy risks.
Further, recent work has shown that general purpose large language models are prone to memorizing sensitive information which can subsequently be extracted \cite{Carlini2020ExtractingTD}.
In the context of biomedical NLP, such concerns have been cited as reasons for withholding
direct publication of trained model weights \cite{mckinney2020reply}.
These uncertainties will continue to hamper dissemination
of trained models among the broader biomedical NLP research community, motivating a need to investigate the susceptibility of such models to adversarial attacks. 

This work is a first step towards exploring the potential privacy implications of sharing model weights induced over non-deidentified EHR text.
We propose and run a battery of experiments intended to evaluate the degree to which Transformers (here, BERT) pretrained via standard masked language modeling objectives over notes in EHR might reveal sensitive information (Figure \ref{fig:overview}).\footnote{We consider BERT rather than an auto-regressive language model such as GPT-* given the comparatively widespread adoption of the former for biomedical NLP.}

We find that simple methods are able to recover associations between patients and conditions at rates better than chance, but not with performance beyond that achievable using baseline condition frequencies.
This holds even when we enrich clinical notes by explicitly inserting patient names into every sentence.
Our results using a recently proposed, more sophisticated attack based on \emph{generating} text \cite{Carlini2020ExtractingTD} are mixed, and constitute a promising direction for future work.





\section{Related Work}

Unintended memorization by machine learning models has significant privacy implications, especially where models are trained over non-deidentified data.
\citet{Carlini2020ExtractingTD} was recently able to extract memorized content from GPT-2 with up to 67\% precision. 
This raises questions about the risks of sharing parameters of models trained over non-deidentified data. 
While one may mitigate concerns by attempting to remove PHI from datasets, no approach will be perfect \cite{BeaulieuJones2018PrivacyPreservingDD,deid_Johnson}. 
Further, deidentifying EHR data is a laborious step that one may be inclined to skip for models intended for internal use.
An important practical question arises in such situations: Is it safe to share the trained model parameters? 

While prior work has investigated issues at the intersection of neural networks and privacy \cite{Song2018TheNA,Salem2019MLLeaksMA,Fredrikson2015ModelIA}, we are unaware of work that specifically focuses on attacking the modern Transformer encoders widely used in NLP (e.g., BERT) trained on EHR notes, an increasingly popular approach in the biomedical NLP community.
In a related effort, \citet{word_embedding_privacy} explored the risks of using imperfect deidentification algorithms together with \emph{static} word embeddings,
finding that such embeddings do reveal sensitive information to at least some degree.
However, it is not clear to what extent this finding holds for the \emph{contextualized} embeddings induced by large Transformer architectures. 

Prior efforts have also applied template and probe-based methods \cite{Bouraoui2020InducingRK,Petroni2019LanguageMA,Zhengbao2020LM,Roberts2020HowMK,Heinzerling2020LanguageMA} to extract relational knowledge from large pretrained models; we draw upon these techniques in this work. 
However, these works focus on general domain knowledge extraction, rather than clinical tasks which pose unique privacy concerns.


\section{Dataset}
\label{sec:dataset}
We use the Medical Information Mart for Intensive Care III (MIMIC-III) English dataset to conduct our experiments \cite{mimiciii}.
We follow prior work \cite{Huang2019ClinicalBERTMC} and remove all notes except for those categorized as `Physician', `Nursing', `Nursing/Others', or `Discharge Summary' note types.
The MIMIC-III database was deidentified using a combination of regular expressions and human oversight, successfully removing almost all forms of PHI \cite{automatic_deid_2008}.
All patient first and last names were replaced with {\tt [Known First Name ...]} and {\tt [Known Last Name ...]} pseudo-tokens respectively. 

We are interested in quantifying the risks of releasing contextualized embedding weights trained on \emph{non-deidentified} text (to which one working at hospitals would readily have access).
To simulate the existence of PHI in the MIMIC-III set, we randomly select new names for all patients \cite{Stubbs2015ChallengesIS}.\footnote{We could have used non-deidentified EHRs from a hospital, but this would preclude releasing the data, hindering reproducibility.} 
Specifically, we replaced {\tt [Known First Name]} and {\tt [Known Last Name]} with names sampled from US Census data, randomly sampling first names (that appear at least 10 times in census data) and last names (that appear at least 400 times).\footnote{We sampled first and last names from \textsf{\small \url{https://www.ssa.gov/}} and \textsf{\small \url{https://www.census.gov/topics/population/genealogy/data/2010\_surnames.html}}, respectively.}


This procedure resulted in 11.5\% and 100\% of patients being assigned unique first and last names, respectively.
While there are many forms of PHI, we are primarily interested in recovering name and condition pairs, as the ability to infer with some certainty the specific conditions that a patient has is a key privacy concern.
This is also consistent with prior work on static word embeddings learned from EHR \cite{word_embedding_privacy}.

Notes in MIMIC-III do not consistently explicitly reference patient names.
First or last names are mentioned in at least one note for only 27,906 (out of 46,520) unique patients.\footnote{In some sense this bodes well for privacy concerns, given that language models are unlikely to memorize names that they are not exposed to; however, it is unclear how particular this observation is to the MIMIC corpus.}
Given that we cannot reasonably hope to recover information regarding tokens that the model has not observed, in this work we only consider records corresponding to these 27,906 patients.
Despite comprising 61.3\% of the total number of patients, these 27,906 patients are associated with the majority (82.6\%) of all notes (1,247,291 in total).
Further, only 10.2\% of these notes contain at least one mention of a patient's first or last name. 

Of the 1,247,291 notes considered, 17,044 include first name mentions, and 220,782 feature last name mentions. 
Interestingly, for records corresponding to the 27,906 patients, there are an additional 18,345 false positive last name mentions and 29,739 false positive first name mentions; in these cases the name is also an English word (e.g., `young').
As the frequency with which patient names are mentioned explicitly in notes may vary by hospital conventions, we also present semi-synthetic results in which we insert names into notes such that they occur more frequently.

\section{Enumerating Conditions}
\label{section:data-conditions}

As a first attempt to evaluate the risk of BERT leaking sensitive information, we define the following task: Given a patient name that appears in the set of EHR used for pretraining, query the model for the conditions associated with this patient.
Operationally this requires defining a set of conditions against which we can test each patient. 
We consider two general ways of enumerating conditions: (1) Using  International Classification of Diseases, revision 9 (ICD-9) codes attached to records, and (2) Extracting condition strings from the free-text within records.\footnote{In this work, we favor the adversary by considering the set of conditions associated with reidentified patients only.}
Specifically, we experiment with the following variants.
 
\vspace{0.5em}
\noindent \textbf{[ICD-9 Codes]} We collect all ICD-9 codes associated with individual patients.
ICD-9 is a standardized global diagnostic ontology maintained by
the World Health Organization.
Each code is also associated with a description of the condition that it represents. In our set of 27,906 patients, we observe 6,841 unique ICD-9 codes. We additionally use the short ICD-9 code descriptions, which comprise an average of 7.03 word piece tokens per description (under the BERT-Base tokenizer). On average, patient records are associated with 13.6 unique ICD-9 codes.
     
\vspace{0.5em}
\noindent \textbf{[MedCAT]} ICD-9 codes may not accurately reflect patient status, and may not be the ideal means of representing conditions. 
Therefore, we also created lists of conditions to associate with patients by running the MedCAT concept annotation tool \cite{kraljevic2020multidomain} over all patient notes. 
We only keep those extracted entities that correspond to a Disease / Symptom, which we use to normalize condition mentions and map them to their UMLS \cite{Bodenreider2004TheUM} CUI and description. This yields 2,672 unique conditions from the 27,906 patient set. 
On average, patients are associated with an average of 29.5 unique conditions, and conditions comprise 5.37 word piece tokens.
     
 Once we have defined a set of conditions to use for an experiment, we assign binary labels to patients indicating whether or not they are associated with each condition. 
 We then aim to recover the conditions associated with individual patients.


\section{Model and Pretraining Setup}
\subsection{Contextualized Representations (BERT)}

We re-train BERT \cite{Devlin2019BERTPO} over the EHR data described in 
Section \ref{sec:dataset} following the process outlined by \citet{Huang2019ClinicalBERTMC},\footnote{\url{https://github.com/kexinhuang12345/clinicalBERT/blob/master/notebook/pretrain.ipynb}} 
yielding our own version of ClinicalBERT. 
However, we use full-word (rather than wordpiece) masking, due to the performance benefits this provides.\footnote{\url{ https://github.com/google-research/bert}}
We adopt 
hyper-parameters from \citet{Huang2019ClinicalBERTMC}, most importantly using three duplicates of static masking.
We list all model variants considered in Table \ref{tab:bert-setups} (including Base and Large BERT models).
We verify that we can reproduce the results of \citet{Huang2019ClinicalBERTMC} for the 30-day readmission from the discharge summary prediction task.

\begin{table*}[h]
    \centering
    \small
    \begin{tabular}{l|ccc}
    \hline
        Model Name & Starts from & Train iterations (seqlen 128) & Train iterations (seqlen 512) \\\hline
        Regular Base & BERT Base &  300K & 100K \\
        Regular Large & BERT Large & 300K & 100K \\
        Regular Base++ & BERT Base & 1M & - \\ 
        Regular Large++ & BERT Large & 1M & - \\
        Regular Pubmed-base & PubmedBERT-base \cite{pubmedbert} & 1M & - \\\hline
        Name Insertion & BERT base & 300K & 100K \\
        Template Only & BERT base & 300K & 100K \\
        \hline
    \end{tabular}
    \caption{BERT model and training configurations considered in this work. Train iterations are over notes from the MIMIC-III EHR dataset.}
    \label{tab:bert-setups}
\end{table*}

We also consider two \emph{easier} semi-synthetic variants, i.e., where we believe it should be more likely that an adversary could recover sensitive information.
For the \textbf{Name Insertion Model}, we insert (prepend) patient names to \emph{every sentence} within corresponding notes (ignoring grammar), and train a model over this data. Similarly, for the \textbf{Template Only Model}, for each patient and every MedCAT condition they have, we \emph{create} a sentence of the form: ``{\tt [CLS]} Mr./Mrs. {\tt [First Name]} {\tt [Last Name]} is a yo patient with {\tt [Condition]} {\tt [SEP]}". 
This over-representation of names should make it easier to recover information about patients.

\subsection{Static Word Embeddings}
We also explore whether PHI from the MIMIC database can be retrieved using \emph{static} word embeddings derived via CBoW and skip-gram word2vec models \cite{mikolov2013efficient}. 
Here, we follow prior work (\citealt{word_embedding_privacy}; this was conducted on a private set of EHR, rather than MIMIC).
We induce embeddings for (multi-word) patient names and conditions by averaging constituent word representations. We then calculate cosine similarities between these patient and condition embeddings (See Section~\ref{sec:cosine}).

\section{Methods and Results}
We first test the degree to which we are able to retrieve conditions associated with a patient, given their name. 
(We later also consider a simpler task: Querying the model as to whether or not it observed a particular patient name during training.)
All results presented are derived over the set of 27,906 patients described in Section \ref{section:data-conditions}. 

The following methods output scalars indicating the likelihood of a condition, given a patient name and learned BERT weights.
We compute metrics with these scores for each patient, measuring our ability to recover patient/condition associations. 
We aggregate metrics by averaging over all patients.
We report AUCs and \emph{accuracy at 10} (A@10), i.e., the fraction of the top-10 scoring conditions that the patient indeed has (according to the reference set of conditions for said patient). 

\subsection{Fill-in-the-Blank}

We attempt to reveal information memorized during pretraining using masked \emph{template} strings. 
The idea is to run such templates through BERT, and observe the rankings induced over conditions (or names).\footnote{This is similar to methods used in work on evaluating language models as knowledge bases \cite{Petroni2019LanguageMA}.}
This requires specifying templates.

\begin{table}
\small
\centering
\begin{tabular}{lll}
\hline
Model & AUC & A@10\\
\hline
\textbf{ICD9} \\
Frequency Baseline &  \textbf{0.926} & 0.134\\
Regular Base & 0.614 & 0.056 \\
Regular Large & 0.654 & 0.063 \\
Name Insertion & 0.616 & 0.057 \\
Template Only & 0.614 & 0.050 \\
\hline
\textbf{MedCAT} \\
Frequency Baseline &  \textbf{0.933} & 0.241\\
Regular Base & 0.529 & 0.109 \\
Regular Large & 0.667 & 0.108 \\
Name Insertion & 0.541 & 0.112 \\
Template Only & \textbf{0.784} & 0.160 \\
\hline
\end{tabular}
\caption{Fill-in-the-Blank AUC and accuracy at 10 (A@10). The Frequency Baseline ranks conditions by their empirical frequencies.  
Results for Base++, Large++, Pubmed-Base models are provided in Appendix Table \ref{table:predict_condition_given_name_spearman}.}
\label{table:icd_medcat_regular_performance}
\end{table}

\paragraph{Generic Templates} We query the model to fill in the masked tokens in the following sequence: ``{\tt [CLS]} Mr./Mrs. {\tt [First Name]} {\tt [Last Name]} is a yo patient with {\tt [MASK]$^+$} {\tt [SEP]}". Here, Mr. and Mrs. are selected according to the gender of the patient as specified in the MIMIC corpus.\footnote{We do not include age as \citet{Huang2019ClinicalBERTMC} does not include digits in pretraining.}
The {\tt [MASK]$^+$} above is actually a sequence of {\tt [MASK]} tokens, where the length of this sequence depends on the length of the tokenized condition for which we are probing.

Given a patient name and condition, we compute the perplexity (PPL) for condition tokens as candidates to fill the template mask. 
For example, if we wanted to know whether a patient (``John Doe") was associated with a particular condition (``MRSA"), we would query the model with the following (populated) template: ``{\tt [CLS]} Mr. John Doe is a yo patient with {\tt [MASK]} {\tt [SEP]}" 
and measure the perplexity of ``MRSA'' assuming the {\tt [MASK]} input token position.
For multi-word conditions, we first considered taking an average PPL over constituent words, but this led to counterintuitive results: longer conditions tend to yield lower PPL. In general, multi-word targets are difficult to assess as PPL is not well-defined for masked language models like BERT \cite{jiang20emnlp,salazar-etal-2020-masked}. 
Therefore, we bin conditions according to their wordpiece length and compute metrics for bins individually. 
This simplifies our analysis, but makes it difficult for an attacker to aggregate rankings of conditions with different lengths.

\paragraph{Results} We use the generic template method to score ICD-9 or MedCAT condition descriptions for each patient. 
We report the performance (averaged across length bins) achieved by this method in Table \ref{table:icd_medcat_regular_performance}, with respect to AUC and A@10.
This straightforward approach fares better than chance, 
but worse than a baseline approach of assigning scores equal to the empirical frequencies of conditions.\footnote{We note that these frequencies are derived from the MIMIC data, which affords an inherent advantage, although it seems likely that condition frequencies derived from other data sources would be similar. We also note that some very common conditions are associated with many patients --- see Appendix Figures \ref{fig:icd9_cat_dist} and \ref{fig:med_cat_dist} --- which may effectively `inflate' the AUCs achieved by the frequency baseline.} 
Perhaps this is unsurprising for MIMIC-III, as only 0.3\% of sentences explicitly mention a patient's last name. 

If patient names appeared more often in the notes, would this approach fare better? 
To test this, we present results for the {\bf Name Insertion} and {\bf Template Only}
variants in Table \ref{table:icd_medcat_regular_performance}. 
Recall that for these we have artificially increased the number of patient names that occur in the training data; this should make it easier to link conditions to names.
The {\bf Template Only} variant yields better performance for MedCAT labels, but still fares worse than ranking conditions according to empirical frequencies.
However, it may be that the frequency baseline performs so well simply due to many patients sharing a few dominating conditions. 
To account for this, we additionally calculate performance using the \textbf{Template Only} model on MedCAT conditions that fewer than 50 patients have. We find that the AUC is 0.570, still far lower than the frequency baseline of 0.794 on this restricted condition set.

Other templates, e.g., the most common phrases in the train set that start with a patient name and end with a condition, performed similarly.

\paragraph{Masking the Condition (Only)} 

Given the observed metrics achieved by the `frequency' baseline, we wanted to establish whether models are effectively learning to (poorly) approximate condition frequencies, which might in turn allow for the better than chance AUCs in Table \ref{table:icd_medcat_regular_performance}.
To evaluate the degree to which the model encodes condition frequencies we design a simple template that includes only a masked condition between {\tt [CLS]} and {\tt [SEP]} token (e.g., {\tt [CLS]} {\tt [MASK]}\ldots {\tt [MASK]} {\tt [SEP]}). 
We then calculate the PPL of individual conditions filling these slots. 
In Table \ref{table:predict_only_condition}, we report AUCs, A@10 scores, and Spearman correlations with frequency scores (again, averaged across length bins). The latter are low, suggesting that the model rankings differ from overall frequencies. 

\begin{table}
\small
\centering
\begin{tabular}{lllll}
\hline
Model & AUC & A@10 & Spearman\\
\hline
\textbf{ICD-9} \\
Regular Base & 0.496 & 0.042 &     0.114 \\
Regular Large & 0.560 & 0.049 &     0.109 \\
Name Insertion & 0.483 & 0.042 &     0.100 \\
Template Only & 0.615 & 0.056 &     0.240 \\
\hline
\textbf{MedCAT} \\
        Regular Base & 0.472 & 0.110 &     0.218 \\
       Regular Large & 0.530 & 0.113 &     0.173 \\
      Name Insertion & 0.473 & 0.102 &     0.156 \\
       Template Only & 0.595 & 0.110 &     0.248 \\
\hline
\end{tabular}
\caption{Average AUC, A@10 and Spearman correlations over conditions binned by description length. Correlations are w/r/t empirical condition frequencies.} 
\label{table:predict_only_condition}
\end{table}

\subsection{Probing}

The above token prediction infill setup attacks
the model only via fixed templates.
But the induced representations might implicitly encode
sensitive information that happens to not be readily exposed by the template.
We therefore also investigate a \emph{probing} setup
\cite{alain:iclr17,Bouraoui2020InducingRK}, in which a representation induced by a pretrained model is provided to a second probing model which is trained to predict attributes of interest.
Unlike masked token prediction, probing requires that the adversary have access to a subset of training data to associate targets with representations.

We train an MLP binary classifier on top of the encoded {\tt CLS} token from the last layer of BERT. The probe is trained to differentiate positive instances (conditions the patient has) from negative examples (conditions the patient does \emph{not} have) on a randomly sampled subset of 5000 patients (we downsample the negative class for balancing). We use the following template to encode the patient-condition pairs: ``{\tt [CLS]} Mr./Mrs. {\tt [NAME]} is a patient with {\tt [CONDITION] [SEP]}". For more information on the setup, see Section~\ref{sec:app-allprobing}.
Results are reported in Table \ref{table:probing_experiments}. 
For comparison, we also consider a simpler, ``condition only" template of ``{\tt [CLS] [CONDITION] [SEP]}", which does not include the patient name.

We run experiments on the {\bf Base}, {\bf Large}, and {\bf Name Insertion} models. 
These models achieve strong AUCs, nearly matching the frequency baseline performance in Table \ref{table:icd_medcat_regular_performance}.\footnote{Though the AUCs for the probing are calculated over a randomly sampled test subset of the full data used in Table \ref{table:icd_medcat_regular_performance}.}
However, it appears that removing the patient's name and simply encoding the condition to make a binary prediction yields similar (in fact, slightly better) performance. 
This suggests that the model is mostly learning to approximate condition frequencies.

\begin{table}
\small
\centering
\begin{tabular}{lllll}
\hline
 & \multicolumn{2}{c}{Name + Condition} & \multicolumn{2}{c}{Condition Only} \\
Model & AUC & A@10 & AUC & A@10\\
\hline
\textbf{ICD-9} \\
Standard Base & 0.860 & 0.131 & 0.917 & 0.182 \\
Regular Base &  0.917 & 0.148 &  0.932 & 0.195\\
Regular Large &  0.909 & 0.153  &  0.922 & 0.186 \\
Name Insertion &  0.871 & 0.095 &  0.932 & 0.204\\
\hline
\textbf{MedCAT} \\
Standard Base & 0.918 & 0.355 & 0.954 & 0.464 \\
Regular Base & 0.946 & 0.431 & 0.956 & 0.508\\
Regular Large & 0.942 & 0.393 & 0.955 & 0.475\\
Name Insertion & 0.925 & 0.365 & 0.950 & 0.431\\
\hline
\end{tabular}
\caption{Probing results using BERT-encoded {\tt CLS} tokens on the test set. We use 10,000 patients out of 27,906 due to time constraints. Standard Base is the original BERT base model.} 
\label{table:probing_experiments}
\end{table}

 The standard probing setup encourages the model to use the frequency of target conditions to make predictions. 
 To address this, we also consider a variant in which we probe for only \emph{individual} conditions, rather than defining a single model probing for multiple conditions, as above.
 This means we train independent models per condition, which can then be used to score patients with respect to said conditions.
 To train such models we upsample positive examples such that we train on balanced sets of patients for each condition.\footnote{We upsample the minority examples, rather than undersampling as before, because the single-condition models are comparatively quick to train.}

 This approach provides results for each condition which vary in frequency. 
 To assess the comparative performance of probes over conditions of different prevalence, we group conditions into mutually exclusive bins reflecting frequency (allowing us to analyze differences in performance, e.g., on rare conditions).
 We group conditions by frequencies, from rarest (associated with 2-5 patients) to most common (associated with $>$20 patients).
 We randomly sample 50 conditions from each of these groups, and train an MLP classifier on top of the encoded {\tt CLS} token from the last layer in BERT (this results in 50 \emph{different} models per group, i.e., 200 independent models).
We measure, in terms of AUC and A@10, whether the probe for a condition return comparatively higher scores for patients that have that condition.

We report results in Table \ref{tab:single_condition_probing_experiments}. Except for the rarest conditions (associated with $<$5 patients), these models achieve AUCs that are at best modestly better than chance, with all A@10 metrics $\approx$0. In sum, these models do not meaningfully recover links between patients and conditions.

\begin{table}
\small
\centering
\begin{tabular}{llllll}
\hline
Model & (1,5] & (5,10] & (10,20] & (20, 10k]\\
\hline
\textbf{ICD-9}\\
Regular Base & 0.520 & 0.507 & 0.500 & 0.526\\
Regular Large & 0.444 & 0.505 & 0.479 & 0.522\\
Name Insertion & 0.477 & 0.484 & 0.491 & 0.504\\
\hline
\textbf{MedCAT}\\
Regular Base & 0.481 & 0.534 & 0.525 & 0.487\\
Regular Large & 0.439 & 0.531 & 0.519 & 0.509\\
Name Insertion & 0.460 & 0.577 & 0.508 & 0.525\\
\hline
\end{tabular}
\caption{Probing results (AUCs) for conditions with different frequencies. We make predictions for conditions using independent models based on BERT-encoded {\tt CLS} tokens. We use a 50/50 train/test split over patients (results are over the test set). Columns correspond to conditions of different frequencies, with respect to the number of patients with whom they are associated (headers provide ranges). All A@10 $\approx$ 0.}
\label{tab:single_condition_probing_experiments}
\end{table}

\subsection{Differences in Cosine Similarities}
\label{sec:cosine}

Prior work \cite{word_embedding_privacy} has demonstrated that static word vectors can leak information: The cosine similarities between learned embeddings of patient names and conditions are on average significantly smaller than the similarities between patient names and conditions they do not have. 
We run a similar experiment to investigate whether contextualized embeddings similarly leak information (and also to assess the degree to which this holds on the MIMIC corpus as a point of comparison). 
We calculate the average cosine similarity between learned embeddings of patient names and those of \emph{positive} conditions (conditions that the patient has) minus negative conditions (those that they do not have). 
Conditions and names span multiple tokens; we perform mean pooling over these to induce embeddings. 
Here again we evaluate on the aforementioned set of 27,906 patients. 


We report results for BERT and word2vec (CBoW and SkipGram; \citealt{mikolov2013efficient}) in Table \ref{table:cos_sim_experiments}.\footnote{We provide additional results in the Appendix, including results for alternative pooling strategies and results on the original MIMIC dataset; all yield qualitatively similar results.}
Values greater than zero here suggest leakage, as this implies that patient names end up closer to conditions that patients have, relative to those that they do not.
Even when trained over the {\bf Name Insertion} data (which we manipulated to frequently mention names), we do not observe leakage from the contextualized embeddings. 

\begin{table}
\small
\centering
\begin{tabular}{lrr}
\hline
Model & Mean & Std. \\
\hline
\textbf{ICD-9} \\
Regular Base & -0.010 & 0.019 \\
Regular Large & -0.045 & 0.052 \\
SkipGram Base & 0.004 & 0.050\\ 
CBoW Base & 0.008 & 0.035\\ 
\hline 
BERT Name Insertion &   -0.007 & 0.017 \\
SkipGram Name Insertion & 0.019 & 0.040\\ 
CBoW Name Insertion & 0.017 & 0.043\\ 
\hline
\textbf{MedCAT} \\
Regular Base & -0.037 & 0.015 \\
Regular Large & -0.055 & 0.029 \\
SkipGram Base & -0.011 & 0.024\\ 
CBoW Base & -0.001 & 0.022\\ 
\hline 
BERT Name Insertion & -0.027 & 0.013 \\
SkipGram Name Insertion & 0.013 & 0.024\\ 
CBoW Name Insertion & 0.015 & 0.026\\ 
\hline
\end{tabular}
\caption{Differences in (a) similarities between patient names and conditions they have, and (b) similarities between patient names and conditions they do not have. 
Static embeddings are 200 dimensional; we train these for 10 epochs. For BERT models, we use 10k patients rather than the $\sim$28k due to compute constraints.}
\label{table:cos_sim_experiments}
\end{table}

\subsection{Can we Recover Patient Names?}

Here we try something even more basic: We attempt to
determine whether a pretrained model has seen a particular patient name in training.
The ability to reliably recover 
individual patient names (even if not linked to specific conditions) from BERT models trained over EHR data would be concerning if such models were to be made public.
We consider a number of approaches to this task.

\paragraph{Probing}
We encode the patient's name ({\tt [CLS] [NAME] [SEP]}) using BERT and train a Logistic Regression classifier that consumes resultant {\tt CLS} representations and predicts whether the corresponding patient has been observed in training.

As mentioned above, patient names are explicitly mentioned in notes for 27,906 patients; these constitute our positive examples, and the remaining patients (of the 46,520) are negative examples. 
We split the data into equally sized train and test sets. 
We report results in Table \ref{table:name_probing_experiments}. 
To contextualize these results, we also run this experiment on the standard BERT base model (which is not trained on this EHR data). 
We observe that the AUCs are near chance, 
and that the performance of the standard BERT base model is relatively similar to that of the Regular and Large base models, despite the fact that the standard BERT base model has not seen any notes from MIMIC.


\begin{table}
\small
\centering
\begin{tabular}{llll}
\hline
Model & AUC \\
\hline
Regular Base & 0.508 \\
Large Base & 0.501 \\
\hline
Standard Base & 0.498 \\
\hline
\end{tabular}
\caption{Predictions (on a test set) of which names have been seen by the model. We include the standard BERT \cite{Devlin2019BERTPO} model (``Standard Base"), which is not trained on MIMIC, as a comparator.} 
\label{table:name_probing_experiments}
\end{table}

\subsection{Does observing part of a name reveal more information?}
Given a first name, can we predict whether we have seen a corresponding last name? 
More specifically, we mask out a patient's last name (but not their first) in the template ``{\tt\small [CLS] [First Name] [MASK]$^+$ [SEP]}'' and record the perplexity of the target last name.  
We take as the set of outputs all 46,520 patient names in the corpus. 

We can also flip this experiment, masking only first names. 
This is intuitively quite difficult, as only 10K / 77M sentences (0.013\%) contain both the patient's first and last name. 
This number includes first and last name mentions that are also other English words (e.g. ``young''). 
Results are reported in Table \ref{tab:first_name_given_last}.
We do observe reasonable signal in the semi-synthetic 
{\bf Name Insertion} and {\bf Template Only} variants.

\begin{table}
\small
\centering
\begin{tabular}{ll}
\hline
Model & AUC \\
\hline 
\textbf{First Name Masked}\\
Regular Base & 0.510 \\
Regular Large & 0.506 \\
Name Insertion & 0.562 \\
Template Only & 0.625 \\
\hline 
\textbf{Last Name Masked}\\
Regular Base & 0.503 \\
Regular Large & 0.498 \\
Name Insertion & 0.517 \\
Template Only & 0.733 \\
\hline
\end{tabular}
\caption{We compute the perplexity of the masked parts of names for all 46,520 patients and measure whether the (27,906) reidentified patients receive lower perplexity, compared to remaining patients.}
\vspace{-1em}
\label{tab:first_name_given_last}
\end{table}

\subsection{Text Generation}

\begin{table*}
\small
\centering
\begin{tabular}{lrrrrr}
\hline
Model & Sent. with Name & First Names & Last Names & A@100 & Name + Positive Condition\\
\hline
Standard BERT Base & 84.7\% & 2.16\% & 7.72\% & 0.34 & 12.17\%\\
Regular Base & 47.9\% & 0.94\% & 3.14\% & 0.16 & 23.53\% \\
Name Insertion & 59.6\% & 2.65\% & 4.56\% & 0.84 & 4.17\% \\
\hline
\end{tabular}
\caption{Results over texts generated by the {\bf Base} and {\bf Name Insertion} models. The `Sent. with Name' column is percentage of extracted sentences that contain a name token. The First and Last name columns show what percent of unique names produced are in the MIMIC dataset. After re-ranking all unique names, we report the percentage of top 100 names that belong to a reidentified patient. Finally, The Name + Positive Condition displays what percent of sentences with a patient's name also contain one of their true (MedCAT) conditions.}
\vspace{-1em}
\label{table:carlini_exp}
\end{table*}

Recent work by \citet{Carlini2020ExtractingTD} 
showed that GPT-2 \cite{radford2019language} memorizes training data, and proposed techniques to efficiently recover sensitive information from this model (e.g., email addresses).
They experimented only with large, auto-regressive language models (i.e., GPT-2), but their techniques are sufficiently general for us to use here.
More specifically, to apply their approaches to a BERT-based model\footnote{Which, at least at present, remains the default encoder used in biomedical NLP.} we must be able to sample text from BERT, which is complicated by the fact that it is not a proper (auto-regressive) language model. 
To generate outputs from BERT we therefore followed a method proposed in prior work \cite{wang2019bert}.
This entails treating BERT as a Markov random field language model and using a Gibbs sampling procedure to generate outputs.
We then analyze these outputs from (a) our regular BERT-based model trained on MIMIC; (b) the {\bf Name Insertion}  model, and; (c) a standard BERT Base model \cite{Devlin2019BERTPO}. 
We generate 500k samples from each, each sample consisting of 100 wordpiece tokens.

\paragraph{Comparator Model Perplexity}
\label{sec:cmp-model-perplex}
Following \citet{Carlini2020ExtractingTD}, we attempt to identify which pieces of generated text are most likely to contain memorized names (in this case, from EHR). 
To this end, we examine segments of the text in which the difference in likelihood of our trained BERT model versus the standard BERT-base model \cite{Devlin2019BERTPO} is high. For the samples generated from the standard BERT-base model (not trained on MIMIC), we use our ClinicalBERT model as the comparator.\footnote{Note that this means that even though samples are generated from a model that cannot have memorized anything in the EHR, using a comparator model that was to re-rank these samples \emph{may} effectively reveal information.} 
Using an off-the-shelf NER tagger \cite{spacy}, we identify samples containing name tokens. 

For each sample, we mask name tokens individually and calculate their perplexity under each of the the respective models. We take the difference between these to yield a score (sequences with high likelihood under the trained model and low likelihood according to the general-domain BERT may contain vestiges of training data) and use it to rank our extracted names; we then use this to calculate A@100. 

As expected, the {\bf Name Insertion} model produced more names than the {\bf Base} model, with approximately 60\% of all sentences containing a name (not necessarily in MIMIC). 
Additionally, the A@100 of the {\bf Name Insertion} model substantially outperforms the {\bf Base} model. 
However, when we use spaCy to examine sentences that contain both a condition and a patient's name (of the 27,906), we find that 23.5\% of the time the patient does indeed have a condition produced by the {\bf Base} model. It is unclear to what extent this reflects memorization of concrete patient-condition pairs per se, as opposed to learning more diffused patient-agnostic distributions of conditions in the MIMIC dataset.
The corresponding statistic for the {\bf Name Insertion} variant (4.17\%) may be low because this tends to produce poor quality outputs with many names, but not many conditions.
This is an intriguing result that warrants further research.

However, we caution that these generation experiments are affected by the accuracy of NER taggers used. 
For example, many of the extracted names tend to also be generic words (e.g., `young', `date', `yo', etc.) which may artificially inflate our scores.
In addition, MedCAT sometimes uses abbreviations as conditions, which may also yield `false positives' for conditions.

\section{Limitations}
This work has important limitations.
We have considered only relatively simple ``attacks", based on token in-filling and probing. 
Our preliminary results using the more advanced generation approach (inspired by \citealt{Carlini2020ExtractingTD}) is a promising future direction, although the quality of generation from BERT --- which is not naturally a language model --- may mitigate this. 
This highlights a second limitation: We have only considered BERT, as it is currently the most common choice of pretrained Transformer in the bioNLP community. 
Auto-regressive models such as GPT-2 may be more prone to memorization.
Larger models (e.g., T5 \cite{raffel2019exploring} or GPT-3 \cite{brown2020language}) are also likely to heighten the risk of data leakage if trained over EHR.

Another limitation is that we have only considered the MIMIC-III corpus here, and the style in which notes are written in this dataset --- names appear very infrequently --- likely renders it particularly difficult for BERT to recover implicit associations between patient names and conditions.
We attempted to address this issue with the semi-synthetic {\bf Name Insertion} variant, where we artificially inserted patient names into every sentence; this did not yield qualitatively different results for most experiments. 
Nonetheless, it is possible that experiments on EHR datasets from other hospitals (with different distributions over tokens and names) would change the degree to which one is able to recover PHI.



Finally, these results for BERT may change under different masking strategies --- for example, dynamic masking \cite{liu2019roberta} or choice of tokenizer. 
Both of these may affect memorization and extraction method performance. 

\section{Conclusions}

We have performed an initial investigation into the degree to which large Transformers pretrained over EHR data might reveal sensitive personal health information (PHI).
We ran a battery of experiments in which we attempted to recover such information from BERT model weights estimated over the MIMIC-III dataset (into which we artificially reintroduced patient names, as MIMIC is deidentified). 
Across these experiments, we found that we were mostly unable to meaningfully expose PHI using simple methods.
Moreover, even when we constructed a variant of data in which we prepended patient names to \emph{every sentence} prior to pretraining BERT, we were still unable to recover sensitive information reliably.
Our initial results using more advanced techniques based on generation (\citealt{Carlini2020ExtractingTD}; Table \ref{table:carlini_exp}) are intriguing but inconclusive at present.

Our results certainly do not rule out the possibility that more advanced methods might reveal PHI.
But, these findings do at least suggest that doing so is not trivial. 
To facilitate further research, we make our experimental setup and baseline probing models available: {\small \url{https://github.com/elehman16/exposing_patient_data_release}}.


\section*{Ethical Considerations} 
This work has ethical implications relevant to patient privacy.
HIPAA prohibits the distribution of PHI, for good reason. 
Without this type of privacy law, patient information, for example, could be passed on to a lender and be used to deny a patient's application for mortgages or credit card. 
It is therefore essential that patient information remain private.
This raises an important practical concerning methods in NLP that we have sought to address: Does releasing models pretrained over sensitive data pose a privacy risk? While we were unable to reliably recover PHI in this work, we hope that this effort encourages the community to develop more advanced attacks to probe this potential vulnerability. We would still advise researchers to err on the side of caution and only consider releasing models trained over fully deidentified data (e.g. MIMIC).

\section*{Acknowledgements}

We thank Peter Szolovits for early feedback on a draft of this manuscript, and the anonymous NAACL reviewers for their comments. 

This material is based upon work supported in part by the National Science Foundation under Grant No. 1901117. This Research was also supported with Cloud TPUs from Google's TensorFlow Research Cloud (TFRC).

\bibliographystyle{acl_natbib}
\bibliography{main}

\newpage
\clearpage
\appendix

\section{Appendix}
\label{sec:appendix}

\renewcommand{\thefigure}{A\arabic{figure}}
\setcounter{figure}{0}

\subsection{Training Our BERT Models}
As mentioned previously, we follow most of the hyperparameters stated in \cite{Huang2019ClinicalBERTMC}. 
The code presented in \citet{Huang2019ClinicalBERTMC} accidentally left out all notes under the category `Nursing/Other'; we added these back in, in addition to any notes that fell under the `Discharge Summaries' summary category.
Our dataset consists of approximately 400M words (ignoring wordpieces). 
The number of epochs (following \citealt{Devlin2019BERTPO}) can be calculated as $$\textrm{num\_steps} \cdot \textrm{batch\_size} \cdot \frac{\textrm{tokens\_per\_seq}}{\textrm{total number of tokens}}$$, which at batch size of 128 and sequence length of 128, comes out to 40 epochs if trained for 1M steps (in the ++ models). For standard models, it comes out to 29 epochs. We used cloud TPUs (v2 and v3) to train our models. All experiments are run on a combination of V100, Titan RTX and Quadro RTX 8000 GPUs. 

\subsection{Condition Distribution}
In Appendix Figures \ref{fig:icd9_cat_dist} and \ref{fig:med_cat_dist}, we can see the distribution of ICD-9 and MedCAT conditions across patients. With respect to the ICD-9 codes, there are only 4 conditions that are shared across 10,000+ patients. This number is 32 for MedCAT conditions. 

\begin{figure}
    \centering
    \includegraphics[width=.45\textwidth]{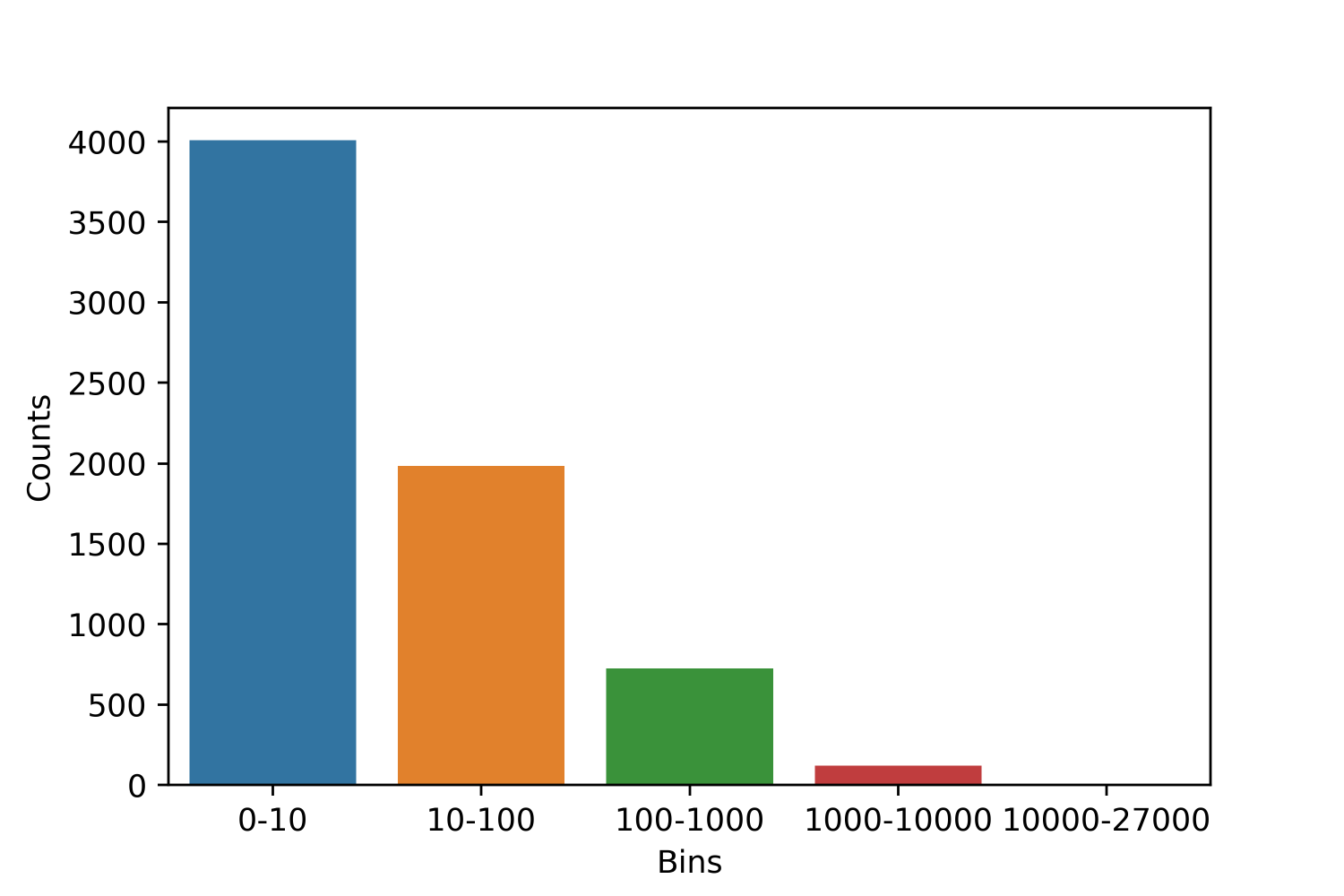}
    \caption{A distribution of ICD-9 codes and how many patients (of the 27K) have each condition. All bin end values are not inclusive.}
    \label{fig:icd9_cat_dist}
\end{figure}

\begin{figure}
    \centering
    \includegraphics[width=.45\textwidth]{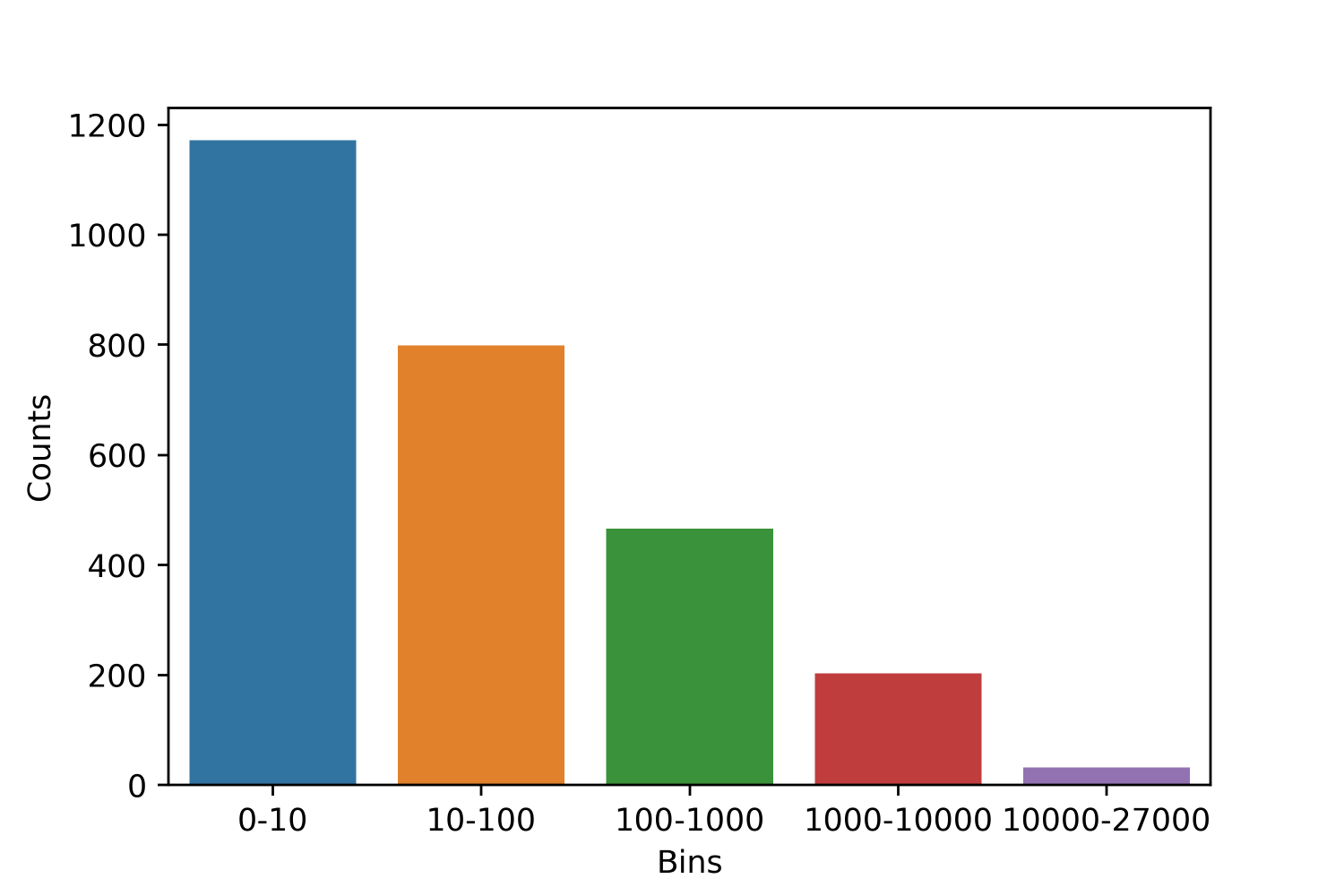}
    \caption{A distribution of MedCAT codes and how many patients (of the 27K) have each condition. All bin end values are not inclusive.}
    \label{fig:med_cat_dist}
\end{figure}

\begin{figure*}%
    \centering
    \subfloat[\centering ICD-9 Labels]{{\includegraphics[width=0.40\textwidth]{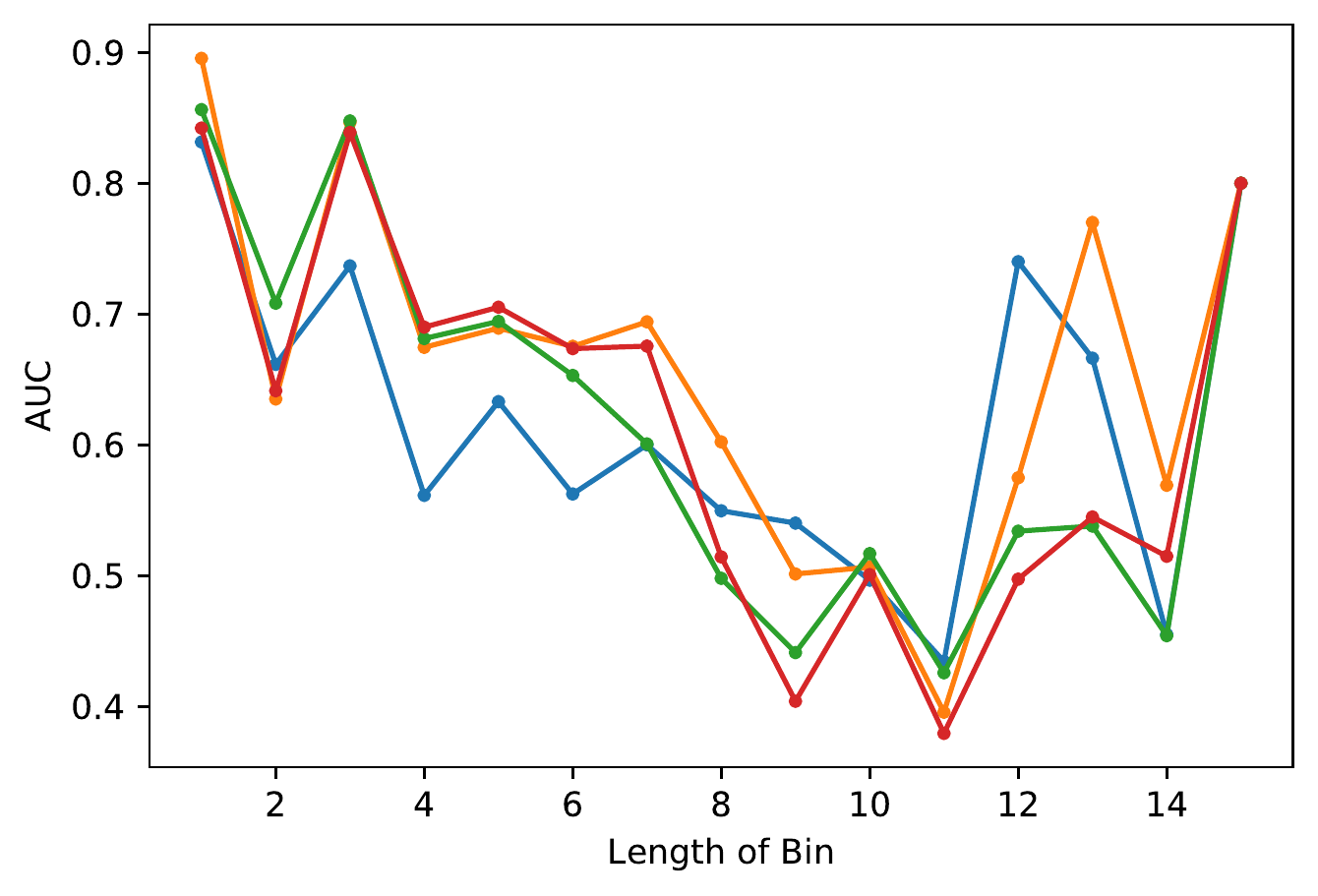} }}%
    \qquad
    \subfloat[\centering MedCAT Labels]{{\includegraphics[width=0.53\textwidth]{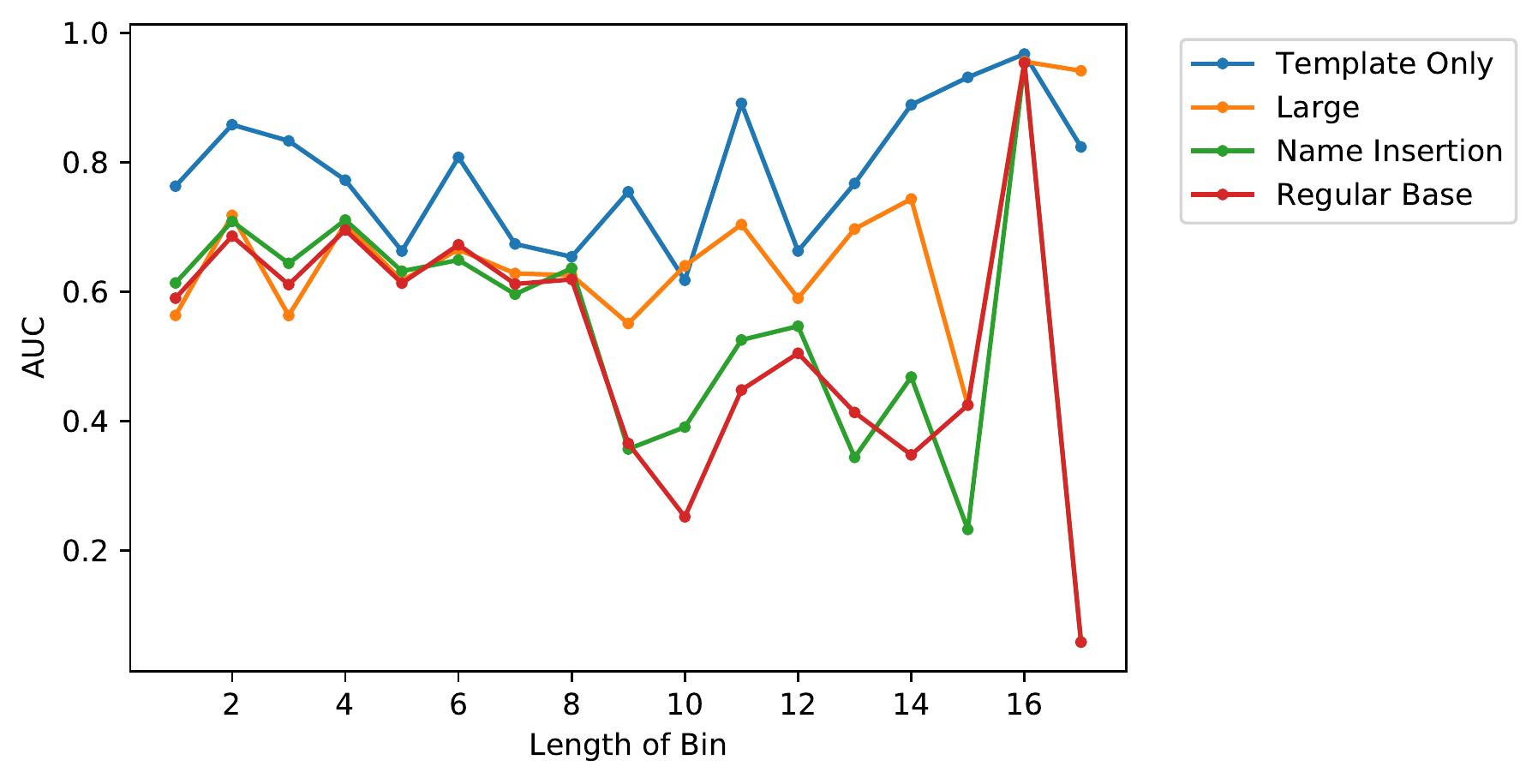} }}%
    \caption{Per-length performance of both ICD-9 and MedCAT labels for the `masked conditon' (only) experiments. A bin length of $k$ contains conditions comprising $k$ token pieces.}
    \label{fig:icd_medcat_per_bin_perf}%
\end{figure*}

\subsection{Condition Given Name}
In addition to the results in Table \ref{table:icd_medcat_regular_performance}, we report all Spearman coefficients, relative to the frequency of conditions (in Appendix Table \ref{table:predict_condition_given_name_spearman}). We additionally report results for Base++, Large++, and Pubmed-Base models. With respect to AUC, these models all perform worse than the Regular Large model. 
Additionally, in Appendix Figure \ref{fig:icd_medcat_per_bin_perf}, we can see how experiment results change with respect to the length of conditions (owing, as we mentioned in the main text, to complications in computing likelihoods of varying length sequences under MLMs).

\begin{table}
\small
\centering
\begin{tabular}{llll}
\hline
Model & AUC & A@10 & Spearman\\
\hline
\textbf{ICD9} \\
        Regular Base & 0.614 & 0.056 &     0.177 \\
       Regular Large & 0.654 & 0.063 &     0.181 \\
      Name Insertion & 0.616 & 0.057 &     0.158 \\
       Template Only & 0.614 & 0.050 &     0.137 \\
      Regular Base++ & 0.588 & 0.059 &     0.141 \\
     Regular Large++ & 0.535 & 0.046 &     0.107 \\
Regular PubmedBase++ & 0.583 & 0.055 &     0.160 \\
\hline
\textbf{MedCAT} \\
        Regular Base & 0.529 & 0.109 &     0.175 \\
       Regular Large & 0.667 & 0.108 &     0.214 \\
      Name Insertion & 0.541 & 0.112 &     0.161 \\
       Template Only & 0.784 & 0.160 &     0.262 \\
      Regular Base++ & 0.511 & 0.109 &     0.124 \\
     Regular Large++ & 0.469 & 0.098 &     0.152 \\
Regular PubmedBase++ & 0.592 & 0.076 &     0.211 \\
\hline
\end{tabular}
\caption{AUC, accuracy at 10 (A@10), and Spearman coefficient relative to condition frequency.}
\label{table:predict_condition_given_name_spearman}
\end{table}

\subsection{Condition Only}
In addition to the results in Table \ref{table:predict_only_condition}, we show results for Base++, Large++, and Pubmed-Base models. Interestingly, the Large and Pubmed-Base model's perform better when names are \emph{not} included. We see the biggest difference between Appendix Table \ref{table:predict_condition_given_name_spearman} and \ref{table:predict_condition_spearman} in the {\bf Templates Only} model, suggesting that this model is memorizing the relationship between patients and conditions.

\begin{table}
\small
\centering
\begin{tabular}{llll}
\hline
Model & AUC & A@10 & Spearman\\
\hline
\textbf{ICD-9} \\
      Regular Base++ & 0.498 & 0.044 &     0.113 \\
     Regular Large++ & 0.516 & 0.044 &     0.076 \\
Regular PubmedBase++ & 0.544 & 0.043 &     0.123 \\
\hline
\textbf{MedCAT} \\
      Regular Base++ & 0.456 & 0.103 &     0.157 \\
     Regular Large++ & 0.454 & 0.113 &     0.122 \\
Regular PubmedBase++ & 0.628 & 0.080 &     0.213 \\
\hline
\end{tabular}
\caption{AUC and A@10 measures with models given only a masked out condition. We calculate spearman coefficients are given relative to the frequency baseline.}
\label{table:predict_condition_spearman}

\end{table}

\subsection{MLP Probing for Names and Conditions}
\label{sec:app-allprobing}

In this experiment, we randomly sample 10,000 patients from our 27,906 patient set (due to computational constraints), of which we keep 5,000 for training and 5,000 for testing. For each of these patient names and every condition in our universe of conditions, we construct the previously specified template and assign it a binary label indicating whether the patient have that condition or not. Since the negative class is over-represented by a large amount in this \emph{training set}, we use downsampling to balance our data. We map each of these templates to their corresponding {\tt CLS} token embedding. We use the embeddings for templates associated with training set patients to train a MLP classifier implemented in Scikit-Learn \cite{scikit-learn} (Note we did not use on a validation set here). We used a hidden layer size of 128 with default hyperparameters. 

At test time, for each of the 5000 patients in test set and each condition, we calculate the score using this MLP probe and compute our metrics with respect to the true label associated with that patient-condition pair.

\subsection{Probing for Individual Conditions}

In this experiment, we samples 50 conditions from each of the 4 \emph{frequency} bins. For each condition, we trained a probe to distinguish between patients that have that condition vs those that do not. This experiment differs from 
the preceding fill-in-the-blank and probing experiments: Here we compute an AUC for each \emph{condition} (indicating whether the probe discriminates between patients that have a particular condition and those that do not),
whereas in the fill-in-the-blank experiments we computed AUCs per \emph{patient}.

For probing individual conditions, we used an MLP classifier implemented in Scikit-Learn \cite{scikit-learn}. We did not evaluate on a validation set. We used a hidden layer size of 128 with default hyperparameters. All experiments were only run once. For the Regular BERT model, we additionally experimented with backpropagating through the BERT weights, but found that this made no difference in predictive performance.

\subsection{Cosine Similarities}
All versions of Skipgram and CBoW \cite{mikolov2013efficient} were trained for 10 epochs using gensim library \cite{rehurek_lrec}, used a vector size of 200, and a window size of 6. We only trained one variant of each W2V model. For BERT models, we used the last layer wordpiece embeddings. For word embedding models, we ran this experiment on whole reidentified patient set, whereas for BERT models, we sampled 10K patients. We report averages over the patients. In addition to the mean-pool collapsing of conditions, we also try `Max-Pooling' and a variant we label as `All Pairs Pooling'. We present results from all cosine-similarity experiments in Appendix Table \ref{table:cosine_similarity_experiment_all_results}. The mean pooling results in Table \ref{table:cos_sim_experiments} seem to outperform the alternative pooling mechanisms presented here. 

\begin{table}
\small
\centering
\begin{tabular}{lrr}
\hline
Model & Mean & Std. \\
\hline
\multicolumn{3}{c}{\textbf{ICD9}} \\
\textbf{Max Pooling} \\
Regular Base &   -0.0093 & 0.017 \\
Regular Large &   -0.020 & 0.029 \\
SkipGram Base &   -0.004 & 0.039 \\ 
CBoW Base &   -0.009 & 0.051\\ 
\hline
Name Insertion &   -0.008 & 0.018 \\
SkipGram Name Insertion &   0.004 & 0.038\\ 
CBoW Name Insertion &   -0.009 & 0.058\\ 
\hline 
\textbf{All Pairs Pooling} \\
Regular Base &   -0.006 & 0.014\\
Regular Large &   -0.029 & 0.042 \\
SkipGram Base &   0.006 & 0.044 \\ 
CBoW Base &   0.005 & 0.044\\ 
\hline 
Name Insertion &   -0.001 & 0.013 \\
SkipGram Name Insertion &   0.019 & 0.039\\ 
CBoW Name Insertion &   0.010 & 0.036\\ 
\hline
\multicolumn{3}{c}{\textbf{MedCAT}} \\
\textbf{Max Pooling} \\
Regular Base &   -0.065 & 0.030 \\
Regular Large &   -0.092 & 0.033 \\
SkipGram Base &   -0.032 & 0.039\\ 
CBoW Base &   -0.071 & 0.059\\ 
\hline 
Name Insertion &   -0.070 & 0.030 \\
SkipGram Name Insertion &   -0.021 & 0.035\\ 
CBoW Name Insertion &   -0.087 & 0.059\\ 
\hline 
\textbf{All Pairs Pooling} \\
Regular Base &   -0.012 & 0.012 \\
Regular Large &   -0.043 & 0.028 \\
SkipGram Base &   -0.005 & 0.020\\ 
CBoW Base &   -0.012 & 0.020\\ 
\hline 
Name Insertion &   -0.011 & 0.009 \\
SkipGram Name Insertion &   0.015 & 0.026\\ 
CBoW Name Insertion &   0.004 & 0.024\\ 
\hline 
\hline
\end{tabular}
\caption{Similarity for Positive Conditions - Negative Conditions. All experiments are performed using ICD-9 codes. Max and Average refer to max-pooling and average-pooling over multiple embeddings, respectively. 
``All" entails the following: For every word piece in the name, find the cosine similarity for every word piece in the condition; then, use the largest cosine similarity. All word embedding models are trained for 10 epochs, with dimensionality 200.}
\label{table:cosine_similarity_experiment_all_results}
\end{table}

\subsection{Probing for Names}
To see if our BERT models are able to recognize the patient names that appear in training data, we train a linear probe on top of names encoded via BERT. We train this Linear Regression classifier using all default parameters from Scikit-Learn (10,000 max steps) \citep{scikit-learn}. We did not evaluate on a validation set. Each experiment was only run once.

\subsection{Does observing part of a name reveal more information?}
Similar to the results in Table \ref{tab:first_name_given_last}, we report results on the Base++, Large++, and Pubmed-Base models (Appendix Table \ref{tab:first_name_given_last_extra_model}). We find no significant difference between these results and the ones reported in Table \ref{tab:first_name_given_last}.

\begin{table}
\small
\centering
\begin{tabular}{ll}
\hline
Model & AUC \\
\hline 
\textbf{First Name}\\
Regular Base++ & 0.505 \\
Regular Large++ & 0.502 \\
Regular Pubmed-base & 0.501 \\
\hline 
\textbf{Last Name}\\
Regular Base++ & 0.504 \\
Regular Large++ & 0.502 \\
Regular Pubmed-base & 0.504\\
\hline
\end{tabular}
\caption{We compute the perplexity of the masked parts of names for all 46,520 patients and measure whether the (27,906) reidentified patients receive lower perplexity, compared to remaining patients.}
\label{tab:first_name_given_last_extra_model}
\end{table}



\end{document}